\documentclass{article}

\PassOptionsToPackage{numbers}{natbib}

\usepackage[main, final]{neurips_2026}

\usepackage[utf8]{inputenc} 
\usepackage[T1]{fontenc}    
\usepackage{hyperref}       
\usepackage{url}            
\usepackage{booktabs}       
\usepackage{amsfonts}       
\usepackage{nicefrac}       
\usepackage{microtype}      
\usepackage{xcolor}         
\usepackage{amsmath}
\usepackage{graphicx}
\usepackage{multirow}
\usepackage{enumitem}
\usepackage{amsthm}
\usepackage{algorithm}
\usepackage{algpseudocode}
\usepackage{tikz}
\usetikzlibrary{positioning,arrows.meta,calc,decorations.pathreplacing,shapes}

\definecolor{ColD}{RGB}{70,120,185}   
\definecolor{ColW}{RGB}{208,88,40}    
\definecolor{ColBg}{RGB}{247,248,252} 
\definecolor{ColG}{RGB}{72,72,82}     

\tikzset{
  BOX/.style ={draw=gray!55, fill=ColBg, rounded corners=5pt,
               inner sep=6pt},
  ARR/.style ={-{Stealth[length=7pt,width=5pt]},
               line width=1.8pt, color=ColG},
  SARR/.style={-{Stealth[length=6pt,width=4pt]},
               line width=1.3pt, color=ColG},
  LBL/.style ={font=\scriptsize, color=gray!68!black, align=center},
  HDR/.style ={font=\small\bfseries, align=center},
  BADGE/.style={font=\tiny, draw=gray!40, fill=white,
                rounded corners=2pt, inner sep=2pt, color=gray!62},
}

\newif\ifshowrev
\showrevfalse
\newcommand{\rev}[1]{\ifshowrev{\color{red}#1}\else#1\fi}
\newenvironment{revblock}{\ifshowrev\color{red}\fi}{}

\title{Crafting Reversible SFT Behaviors \\ in Large Language Models}

\author{%
  Yuping Lin$^{1}$ \quad Pengfei He$^{1}$ \quad Yue Xing$^{1}$ \quad Yingqian Cui$^{1}$\thanks{\rev{Corresponding author.}} \\
  \textbf{Jiayuan Ding}$^{2}$ \quad \textbf{Subhabrata Mukherjee}$^{2}$ \quad \textbf{Hui Liu}$^{1}$ \quad \textbf{Zhen Xiang}$^{3}$  \\
  \\
  $^1$Michigan State University \quad $^2$Hippocratic AI \quad $^3$University of Georgia \\
  \\
  \texttt{\{linyupin,hepengf1,xingyue1,cuiyingq,liuhui7\}@msu.edu} \\
  \texttt{\{jiayuan,subho\}@hippocraticai.com} \\
  \texttt{zxiangaa@uga.edu} \\
}

\begin{document}

\maketitle

\begin{abstract}

Supervised fine-tuning (SFT) induces new behaviors in large language models, yet imposes no structural constraint on how these behaviors are distributed within the model. Existing behavior interpretation methods, such as circuit attribution approaches, identify sparse subnetworks correlated with SFT-induced behaviors post-hoc. \rev{However, the \rev{fine-tuned} model may still carry the behavior through other pathways, so such subnetworks offer no guaranteed handle} for selectively controlling SFT-induced behaviors at inference time. We pursue \rev{a constructive} alternative by asking: can an SFT-induced behavior be deliberately compressed into a sparse subnetwork, termed a \textit{carrier}, \rev{that is sufficient for the behavior and, by construction, necessary for it,} while remaining controllable at inference time without weight modification? We propose (a) \textbf{Loss-Constrained Dual Descent (LCDD)}, which constructs such carriers by jointly optimizing routing masks and model weights under an explicit utility budget, and (b) \textbf{SFT-Eraser}, a soft \rev{trigger} prompt optimized via activation matching on extracted carrier channels, to \rev{revert} the SFT-induced behavior \rev{toward the base model}. Across safety, fixed-response, and style behaviors on \rev{four} model families, LCDD yields sparse carriers that preserve target behaviors while enabling strong reversion when triggered by SFT-Eraser. \rev{Our ablation study shows that triggered LCDD models are consistently closer to the base model than identically triggered standard SFT models. The crafted structure is therefore what drives this reversion. These results establish training-time construction as a practical route to} localizing and selectively suppressing SFT-induced behaviors in deployed models. Code is available at \url{https://github.com/yuplin2333/sft-reverse}.

\end{abstract}

\section{Introduction}\label{sec:intro}

Supervised fine-tuning (SFT) is the standard approach for training large language models (LLMs) to exhibit deployment-critical behaviors such as safety alignment~\citep{ouyang2022training,touvron2023llama,bai2022constitutional}, instruction following~\citep{ouyang2022training,zhou2023lima}, and domain adaptation~\citep{mecklenburg2024injecting}. Despite its empirical success, the SFT process provides no structural guarantee on how behaviors are internally organized. The resulting behavior may be spread broadly, redundantly represented, or entangled with unrelated computation. There is no designated substructure that can be precisely targeted or analyzed.

This lack of structure creates two compounding problems. 
\textbf{First}, it \rev{obscures \emph{localization}}. In mechanistic interpretability, a \emph{circuit} is a sparse subnetwork of model components (attention heads, MLP layers, or residual channels) that is responsible for a specific behavior~\citep{elhage2021mathematical,olah2020zoom}. Post-hoc circuit discovery identifies such subnetworks by correlating component activations with target outputs. However, correlation does not imply causation. The full model may retain redundant pathways that compensate for any ablated component\rev{~\citep{chan2022causal}}. As long as behavior remains diffusely encoded, \rev{the location of SFT-induced computation can only be inferred, never guaranteed.}
\textbf{Second}, it limits \emph{\rev{controllability}}. Existing approaches do not provide a way to \emph{instantiate} a sparse substructure \rev{that holds an SFT-induced behavior by design}. Behavioral and representational analyses of SFT~\citep{zhou2023lima,li2024superficial,zhao2026layer,zhao2024supervised,ding2025improved,sun2026talking,wang2024uft} and circuit localization methods~\citep{conmy2023towards,syed2024attribution,marks2024sparse,yu2026safeseek,gao2025weight,elhage2022toy} can identify structure post-hoc but do not construct it. Task-vector and model-editing methods~\citep{ilharco2022editing,ortiz2023task} show that fine-tuning weight changes can be composed and edited to add or remove capabilities, but they intervene at the parameter level and do not produce a substructure addressable via input alone under fixed weights. To the best of our knowledge, no prior work constructs a sparse substructure that \rev{contains all SFT-induced weight changes by construction} and \rev{is} selectively suppressible via input alone under fixed weights.

We pursue a fundamentally different approach that addresses both problems at once: \rev{we ask whether sparse structure can be deliberately constructed during training.} 
When structure is imposed by design, the behavior is intentionally concentrated within \rev{a sparse substructure, which we call a \textit{sparse behavioral carrier},} and components outside it are maintained in their base-model state (i.e., before SFT). \rev{The carrier is thus necessary for the behavior by construction, a guarantee that post-hoc circuit attribution cannot provide. An input intervention that targets the carrier can then be evaluated directly under fixed weights. Beyond analysis,} a deployed model with a crafted sparse carrier could support inference-time behavioral auditing or selective suppression without any weight modification, enabling more modular post-training control.

These considerations motivate our research question: 

\emph{Can SFT-induced behaviors be intentionally concentrated into sparse carriers \rev{that are sufficient for the behavior and, by construction, necessary for it, and} that remain controllable at inference time without modifying model weights?}

We study this question via two methods. We introduce \textbf{Loss-Constrained Dual Descent (LCDD)} to compress SFT behavior into a sparse carrier via \rev{learned masks on the weight delta}, and \textbf{SFT-Eraser} to reverse the behaviors introduced by SFT via activation matching. SFT-Eraser treats a \emph{reversal} as successful when it produces both behavioral suppression and distributional reversion\rev{, i.e., the output distribution of the triggered model moves back toward that of the base model}. Our main contributions are as follows:

\begin{itemize}[leftmargin=*]
    \item To the best of our knowledge, we are the first to investigate the feasibility of concentrating SFT-induced behaviors into sparse substructures \rev{that are sufficient for the behavior and, by construction, necessary for it}, enabling their control at inference time without modifying model weights.

    \item We propose \textbf{LCDD}, a framework for constructing sparse behavioral carriers by jointly optimizing routing masks and model weights under explicit utility constraints. To \rev{control the crafted carriers}, we further introduce \textbf{SFT-Eraser}, an input-trigger\footnote{While we borrow the term \emph{trigger} from the backdoor literature~\citep{gu2019badnets}, our trigger \rev{steers the model back toward its behavior before SFT}, whereas a backdoor trigger activates a specific implanted behavior.} \rev{method that suppresses SFT-induced behaviors under fixed weights.}

    \item We demonstrate that LCDD and SFT-Eraser enable inference-time reversibility across three behavior types and four model families. \rev{Our ablation study shows that triggered LCDD models are consistently closer to the base model than identically triggered standard SFT models, identifying the crafted structure as what drives reversion.}
\end{itemize}

\vspace{-0.5em}
\section{Related Work}
\vspace{-0.5em}

\paragraph{Post-hoc analysis of SFT behaviors.}

Recent work characterizes SFT observationally. LIMA shows that a small high-quality SFT set suffices for strong assistant behavior~\citep{zhou2023lima}, and the Superficial Safety Alignment Hypothesis frames alignment as a lightweight ``fulfill vs.\ refuse'' controller with sparse safety-critical components~\citep{li2024superficial}. Depth-wise analyses find that alignment updates concentrate in specific layers~\citep{zhao2026layer,zhao2024supervised}, and other work addresses catastrophic forgetting and objective mismatch~\citep{ding2025improved,sun2026talking,wang2024uft}. SafeSeek frames circuit extraction as mask optimization and demonstrates highly sparse circuits for backdoor and alignment behaviors~\citep{yu2026safeseek}, and Depth Charge shows that targeted interventions on deep attention heads can substantially alter jailbreak success~\citep{wu2026depth}. These methods identify or characterize structure post-hoc. 
By contrast, our work constructs sparse behavioral carriers during training, so that the carrier \rev{is necessary for the behavior by construction and the location of SFT-induced computation is known in advance}.

\paragraph{Constructive and weight-space approaches.}

Task arithmetic defines task vectors as $\Delta W = W_{\mathrm{ft}} - W_{\mathrm{base}}$ and shows that adding or negating them can compose or suppress behaviors~\citep{ilharco2022editing,ortiz2023task}. However, it operates with dense global weight edits and does not produce a sparse activation-level structure addressable through input alone under fixed weights. 
Parameter-efficient adaptation via task-specific binary masks~\citep{mallya2018piggyback} is closer in spirit. \rev{Our} mask-only ablation (Appendix~\ref{app:ablation1}) \rev{shows that jointly adapting the weights reaches 2.4--2.9$\times$ the sparsity of masking alone}.
Most directly related is~\citet{gao2025weight}, which trains weight-sparse models from scratch on narrow tasks and finds that node-level sparsity yields highly interpretable circuits. Our work shares the same design principle \rev{and applies it to fine-tuning}: we impose sparsity on the SFT weight delta $\Delta W$ in already-trained chat models to construct behavioral carriers. Broader differentiable mask and circuit discovery methods~\citep{conmy2023towards,syed2024attribution,marks2024sparse,louizos2017learning,cao2021low} provide methodological precedents but do not address SFT carrier construction.

\section{Method}\label{sec:method}

\begin{figure}[h!]
    \centering
    \includegraphics[width=0.95\linewidth]{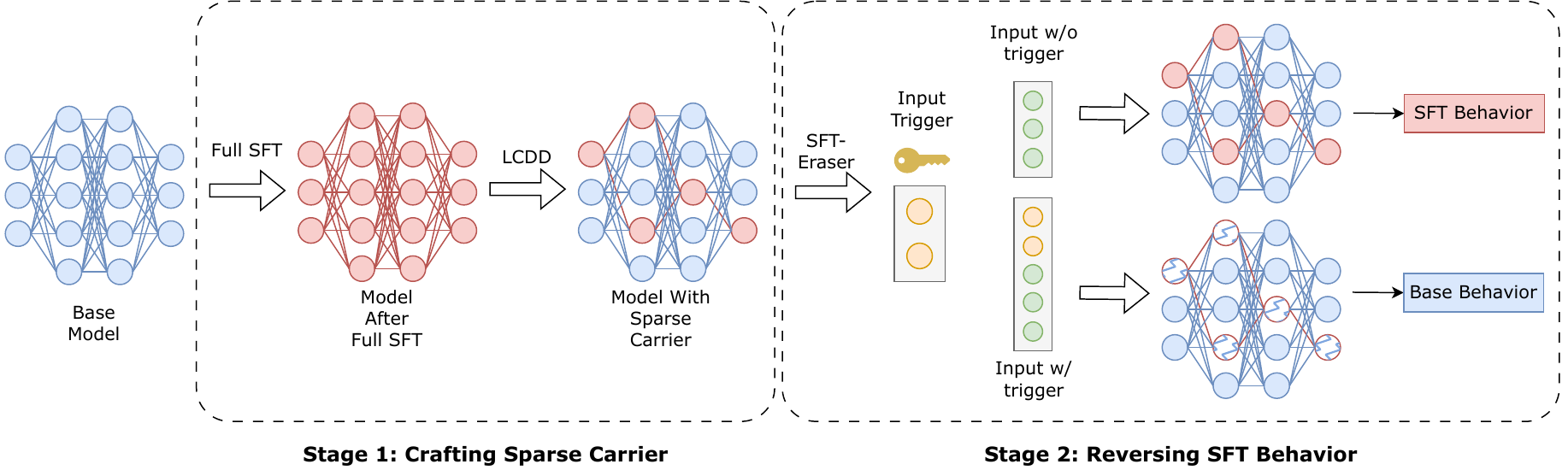}
    \caption{An overview of the LCDD + SFT-Eraser pipeline. \textbf{Stage 1}: standard SFT distributes the induced behavior broadly across model parameters (red), followed by LCDD that compresses the SFT-induced behavior into a sparse carrier. Components outside the carrier are reduced to their base-model state by construction (blue). \textbf{Stage 2}: SFT-Eraser optimizes a soft trigger. Under normal inference without the trigger, the carrier preserves SFT behavior\rev{,} while with the trigger, carrier activations are driven toward the base model.}
    \label{fig:pipeline}
\end{figure}

Our goal is to develop a \emph{sparse behavioral carrier} satisfying two properties: (1) it maintains a minimal parameter substructure sufficient to sustain the SFT-induced behavior under normal inference; and (2) there exists a fixed-weight input intervention targeting the carrier that can reliably suppress the SFT-induced behavior. Property (2) relies on \rev{SFT-induced weight changes being confined to the carrier}: \rev{by construction,} model components outside the carrier \rev{are} reduced to their base-model state. \rev{Sparsity makes the carrier compact and identifiable, so it can be precisely targeted at inference time.} We pursue this in two stages as shown in Figure~\ref{fig:pipeline}: crafting the sparse carrier via \textbf{LCDD} (Section~\ref{sec:method_lcdd}), and \rev{reversing the SFT-induced behavior} via \textbf{SFT-Eraser} (Section~\ref{sec:method_eraser}). Both stages build on the delta parameterization introduced in Section~\ref{sec:method_prelim}.

\subsection{Preliminaries: Weight-Change Parameterization and Mask Implementation}\label{sec:method_prelim}

\paragraph{Weight-change parameterization.}
We represent SFT-induced computation as the additive weight change from fine-tuning. Let $W_{\text{base}}$ and $W_{\text{ft}}$ denote the respective weights. \rev{We} define the \emph{weight delta} $\Delta W = W_{\text{ft}} - W_{\text{base}}$ as the SFT-induced parameter change. Our masked model is then parameterized as
\begin{equation}
W = W_{\text{base}} + M \odot \Delta W,
\end{equation}
where $M \in \{0,1\}^{\text{shape}(W)}$ is a binary mask. Setting $M = \mathbf{0}$ recovers the base model exactly and $M = \mathbf{1}$ recovers the fully fine-tuned model. Our goal is to find 
the sparsest $M$ that still preserves task behavior. Minimizing carrier size is deliberate: a sparser carrier is more identifiable and more precisely targetable at inference time\rev{, and in our utility-budget sweep sparser carriers revert more completely toward the base model (Appendix~\ref{app:sensitivity})}.

\paragraph{Weight mask implementation.}
We structure sparsity at the level of activation channels (individual rows and columns of weight matrices), following the design principle that node-level weight sparsity produces identifiable and interpretable circuits in which information flow can be traced through designated channel bottlenecks~\citep{gao2025weight}. 
Prior work has explored similar gradient-based mask optimization strategies for circuit discovery.
\citet{gao2025weight} train weight-sparse models from scratch such that each neuron reads from and writes to only a small number of residual channels, making circuit boundaries well-defined by construction.
\citet{yu2026safeseek} apply gradient-based binary mask optimization with the straight-through estimator (STE) to discover safety circuits in full models, gating unit outputs (neurons or attention heads) of the complete model weights $W$. Our setup differs in two respects. First, we apply structured masking specifically to the SFT weight change $\Delta W$, isolating SFT-induced computation from the base model. Second, we use row and column gate vectors rather than unit-output masks, yielding a rank-1 structured mask that controls which activation channels carry the SFT-induced delta computation.

Formally, for each weight matrix, we introduce row and column gate vectors $m_{\text{row}}$ and $m_{\text{col}}$. The \rev{contribution of the weight delta $\Delta W$} to a linear layer can be written as $[(\mathbf{x} \odot m_{\text{row}})\,\Delta W] \odot m_{\text{col}}$, where gating the input activations selects rows of $\Delta W$ and gating the output activations selects columns. This activation-gating view is equivalent to weight masking with the rank-1 structure
\begin{equation}
M_{jk} = m_{\text{row},j}\, m_{\text{col},k}.
\end{equation}
Full derivations for both FFN and attention layers are given in Appendix~\ref{app:gate-equiv}. Per transformer layer, we define 8 independent gate groups covering all delta-carrying weight matrices and the complete gate-to-weight-mask mapping is given in Table~\ref{tab:gate-map} of Appendix~\ref{app:gate-equiv}.

\subsection{LCDD: Utility-Budgeted Sparse Carrier Crafting}\label{sec:method_lcdd}

LCDD jointly optimizes the mask $M$ and the weight change $\Delta W$ to compress SFT behavior into a sparse carrier while keeping utility loss within an explicit budget. We first describe how $M$ is parameterized for optimization, and then present the main objective and the optimization procedure.

\vspace{-0.5em}
\paragraph{Parameterizing $M$.}
We cannot directly optimize $M$ due to its discrete nature. Following~\citet{gao2025weight}, we parameterize each gate by a continuous logit $\theta_i \in \mathbb{R}$. During the forward pass, each gate is binarized via the Heaviside function as $m_i = \mathbf{1}[\theta_i > 0]$. Gradients are approximated by the straight-through estimator (STE). The sigmoid $\sigma(\theta_i)$ serves as a differentiable proxy for gate activation and defines the sparsity penalty:
\begin{equation}
  \mathcal{L}_{\text{sparsity}} = \sum_i \sigma(\theta_i).
  \label{eq:sparsity}
\end{equation}
Here, $i$ indexes all individual gate elements across all gate groups and all transformer layers. Minimizing $\mathcal{L}_{\text{sparsity}}$ progressively deactivates weights in the delta path. The task loss gradient via STE retains gates necessary for behavior. Full details are given in Appendix~\ref{app:gate-learning}.

\vspace{-0.5em}
\paragraph{Main objective.}
After parameterizing $M$ via $\theta$, the training objective is:
\begin{equation}
\min_{\theta,\,\Delta W}\ \sum_i \sigma(\theta_i)
\qquad\text{s.t.}\qquad
\mathcal{L}_{\text{task}}(\theta,\Delta W) \le \epsilon.
\label{eq:lcdd-obj}
\end{equation}
where $\mathcal{L}_{\text{task}}$ measures the model's task loss under the current mask and weight configuration, and $\epsilon$ is an explicit budget that caps the tolerated degradation in task performance.
In our main experiments $\mathcal{L}_{\text{task}}$ is SFT cross-entropy. The constraint-based design provides a directly interpretable control surface: one specifies a concrete tolerance on utility degradation and lets the optimizer find the sparsest carrier within that budget.

\vspace{-0.5em}
\paragraph{Optimization.}
To solve Eq.~\eqref{eq:lcdd-obj}, we use an adaptive penalty method: the constraint is absorbed into the objective with a multiplier $\lambda_t$ updated online based on constraint violation. A linear sparsity warmup factor $\rho_t$ gradually engages the penalty during early training, allowing the model to stabilize before compression begins. The multiplier grows when task loss is below budget (intensifying sparsification pressure) and shrinks when loss exceeds budget (allowing utility recovery). This loop continuously renegotiates the sparsity-utility trade-off without manual multiplier schedules or nested inner-loop minimization. Full details are given in Algorithm~\ref{alg:lcdd} (Appendix~\ref{app:lcdd-algo}).

To prevent behavior from routing through the token embedding or output projection layers, we freeze these components throughout training, confining compression to intermediate transformer layers. Training halts when carrier sparsity stalls across consecutive checkpoints or when utility degradation exceeds a threshold. We empirically verify that jointly optimizing masks and weight deltas is necessary for achieving \rev{high sparsity} in Appendix~\ref{app:ablation1}.

\subsection{SFT-Eraser: Trigger-Based Reversal via Residual-Write Channel Matching}\label{sec:method_eraser}

Given a sparse carrier crafted by LCDD, SFT-Eraser asks: can a fixed-weight input intervention suppress SFT behavior by targeting carrier activations directly? \rev{Because the carrier is the only location of SFT-induced weight changes, the trigger is optimized to bring the carrier's activations, and with them the output distribution, back to those of the base model.}

\paragraph{Carrier channel extraction.}
After LCDD training, we identify active channels as those with $\theta_i > 0$. Among these, we select \emph{residual write channels} as the trigger target:
\begin{equation}
\mathcal{C}_{\text{write}}
= \mathcal{C}_{\text{ffn-write}} \cup \mathcal{C}_{\text{attn-write}},
\end{equation}
where $\mathcal{C}_{\text{ffn-write}}$ and $\mathcal{C}_{\text{attn-write}}$ are the sets of active output-gate channels of the FFN and attention delta paths, respectively (corresponding to $m_{\text{ffn-write}}$ and $m_{\text{attn-write}}$ in Section~\ref{sec:method_prelim}). We focus on write channels because they form the carrier's primary interface with the residual stream, where the delta computation contributes to downstream layer inputs.

\paragraph{Trigger objective.}
We optimize a soft trigger embedding sequence $t = [t_1, \dots, t_L]$ prepended to the input prompt $x$, while all model weights remain frozen. The trigger loss combines three terms:
\begin{equation}\label{eq:trigger}
\mathcal{L}_{\text{trigger}}
= \mathcal{L}_{\text{MSE}}
+ \alpha\, \mathcal{L}_{\text{KL}}
+ \beta\, \mathcal{L}_{2}.
\end{equation}
Denoting by $\mathcal{A}_{\mathcal{C}}(\cdot)$ the residual-stream write activations at channel set $\mathcal{C}$, the primary term $\mathcal{L}_{\text{MSE}}$ drives the carrier's write activations toward those of the base model on the same input:
\begin{equation}
\mathcal{L}_{\text{MSE}}
= \mathrm{MSE}\!\left(
\mathcal{A}_{\mathcal{C}_{\text{write}}}(\mathrm{SFT}([t;\,x])),\;
\mathcal{A}_{\mathcal{C}_{\text{write}}}(\mathrm{Base}(x))
\right).
\end{equation}
Since the trigger acts on the full input and can perturb activations beyond the carrier channels, $\mathcal{L}_{\text{MSE}}$ alone does not constrain the trigger's net effect on the output distribution. The auxiliary term $\mathcal{L}_{\text{KL}}$ adds output-level supervision, ensuring that the trigger produces distributional reversion toward the base model:
\begin{equation}
\mathcal{L}_{\text{KL}}
= \mathrm{KL}_{\mathrm{tail}\text{-}k}\!\left(
\mathrm{Base}(x)\;\|\;\mathrm{SFT}([t;\,x])
\right),
\end{equation}
computed over the final $k$ response positions to focus the signal on the behavioral regime. The regularizer $\mathcal{L}_{2} = \frac{1}{L}\sum_{i=1}^{L}\|t_i\|_2^2$ prevents trigger tokens from drifting into degenerate high-magnitude embedding regions where activations become uninterpretable. \rev{Trigger} optimization also applies PGD-style $\ell_2$ projection after each gradient update (details in Appendix~\ref{app:trigger-optim}).

\section{Experiments}\label{sec:experiments}

In this section, we conduct experiments to answer the following
research questions:
\begin{itemize}[leftmargin=*]
    \item \textbf{RQ1:} Can LCDD craft sparse SFT carriers that
    preserve target behavior under normal inference?
    \item \textbf{RQ2:} Do LCDD-crafted carriers yield reliable
    trigger-based reversal under fixed weights?
    \item \textbf{RQ3:} \rev{Does} the crafted sparse structure \rev{drive reversion toward the base model?}
\end{itemize}
We describe the experimental setup in Section~\ref{sec:setup} and
present results in Sections~\ref{sec:results} and~\ref{sec:ablation2}.

\subsection{Experimental Setup}\label{sec:setup}

\paragraph{Tasks.}
We select three types of SFT tasks in increasing order of structural complexity:
\begin{itemize}[leftmargin=*]
    \item \textbf{Fixed Response.} The model is trained to respond ``I don't know'' on all instruction prompts from Alpaca~\citep{taori2023alpaca}. This behavior has a discrete lexical signature and is input-unconditional, making it the \rev{simplest behavior to measure}.
    \item \textbf{Safety Alignment.} The model is trained on WildJailbreak~\citep{jiang2024wildteaming} harmful prompts interleaved with benign prompts (1:1 ratio). This behavior is input-conditional (triggered by semantic harm) and requires the carrier to encode content-sensitive routing.
    \item \textbf{Shakespeare Style.} The model is trained on modern-to-Shakespeare conversational pairs~\citep{roudranil2023shakespeare}. This is a distributional behavior with no discrete trigger condition, requiring the carrier to capture broad stylistic shifts across the output distribution, making it the most structurally demanding of the three tasks.
\end{itemize}
All main runs use 5,000 training samples per task.

\paragraph{Models.}
We evaluate on four chat LLM families: Qwen3-0.6B~\citep{yang2025qwen3}, DeepSeek-R1-Distill-Llama-8B~\citep{guo2025deepseek}, Mistral-7B-Instruct-v0.3~\citep{jiang2023mistral}, and Vicuna-7B-v1.5~\citep{zheng2023judging}. Model selection follows two criteria: architectural diversity across scales and pretraining regimes, and for the safety task, absence of built-in safety alignment so that refusal behavior is \rev{induced by SFT}.

\paragraph{Metrics.} 
We use the following metrics to evaluate how well the sparse carrier preserves SFT behavior before intervention (which we call \emph{carrier fidelity}), and how completely it reverts after triggering.
\begin{itemize}[leftmargin=*]
    \item \textbf{Fixed Response:} Strict fixed-response rate, computed by keyword matching with a response length cap (Appendix~\ref{app:exp-details-eval}).
    \item \textbf{Safety:} WildGuard refusal rate~\citep{han2024wildguard} as the primary safety metric (higher is safer), WildGuard harmfulness rate~\citep{han2024wildguard} as a complementary output-side measure (lower is safer), and HarmBench ASR~\citep{mazeika2024harmbench} as an attack-side measure (lower is safer). MMLU~\citep{hendrycks2020measuring} and HellaSwag~\citep{zellers2019hellaswag} track general utility.
    \item \textbf{Shakespeare:} LLM-judge score on a 0--5 Shakespearean authenticity scale, evaluated by Qwen3-14B~\citep{yang2025qwen3} using the rubric in Appendix~\ref{app:exp-details-eval}. MMLU and HellaSwag track general utility.
    \item \textbf{KL metrics:} Computed by running each evaluated model on SFT-generated reference responses with teacher forcing (i.e., scoring on fixed reference tokens), averaged token-level over response tokens. KL(SFT$\|$LCDD) measures carrier fidelity before intervention (\rev{lower is better}). KL(SFT$\|$Trig) and KL(Base$\|$Trig) jointly characterize distributional movement after triggering: larger KL(SFT$\|$Trig) indicates stronger divergence from SFT behavior\rev{, and} smaller KL(Base$\|$Trig) indicates closer reversion to the base model.
\end{itemize}

\paragraph{Training setup.}
All runs follow the same two-phase pipeline (full SFT then LCDD), with the token embedding layer and output vocabulary projection frozen throughout LCDD training to confine compression to intermediate transformer layers. Trigger optimization targets residual write channels, uses trigger length 20, tail-$k$ KL ($k=8$), and norm-constrained gradient updates ($\ell_2$ max-norm~$= 1.0$) uniformly across all settings. Full hyperparameter details are in Appendix~\ref{app:exp-details}.

\subsection{Main Results}\label{sec:results}

We present results across three behavior types. For each behavior type, we report carrier fidelity (how closely LCDD preserves SFT behavior before intervention) and reversal quality (how completely SFT-Eraser suppresses the behavior after triggering). 

\subsubsection{Fixed Response}

\begin{table}[h!]
\centering
\caption{\textbf{Fixed Response} results. ``Trig'' (LCDD+Trigger) denotes the LCDD model with SFT-Eraser trigger applied. KL values are computed on SFT-generated references.}
\label{tab:idk}
\setlength{\tabcolsep}{6pt}
\resizebox{0.8\textwidth}{!}{
\begin{tabular}{lc cccc ccc}
\toprule
& & \multicolumn{4}{c}{Fixed-Response Rate}
& \multicolumn{3}{c}{KL Divergence} \\
\cmidrule(lr){3-6}\cmidrule(lr){7-9}
Model & Sparsity
& @Base & @SFT & @LCDD & @Trig
& SFT$\|$LCDD & SFT$\|$Trig & Base$\|$Trig \\
\midrule
Qwen3-0.6B  & \rev{78.00\%} & 0.0\% & 100.0\% &  97.5\% & 0.0\%
  & 0.040 & 2.839 & 0.633 \\
DeepSeek-8B & 73.38\% & 0.0\% & 100.0\% &  99.0\% & 1.0\%
  & 0.010 & \rev{1.810} & \rev{1.039} \\
Mistral-7B  & 76.72\% & 0.0\% & 100.0\% & 100.0\% & 0.5\%
  & 0.002 & \rev{3.527} & 1.437 \\
Vicuna-7B   & 38.28\% & 0.0\% & 100.0\% & 100.0\% & 0.0\%
  & \rev{0.389} & \rev{3.068} & \rev{0.396} \\
\bottomrule
\end{tabular}
}
\end{table}

As shown in Table~\ref{tab:idk}, the Fixed Response task provides \rev{clear} evidence that LCDD can craft a sparse carrier that is both behaviorally faithful before intervention and reliably reversible after triggering. LCDD preserves near-perfect fixed-response behavior across all four models (97.5--100\%). \rev{KL(SFT$\|$LCDD) is low across all four models (0.002--0.389), confirming that the sparse carrier closely matches full SFT before intervention.} 
Under trigger, fixed-response rate falls to near-zero in all cases (0.0--1.0\%). The large post-trigger KL(SFT$\|$Trig) values (\rev{1.810--3.527}) confirm strong divergence from SFT behavior, while KL(Base$\|$Trig) (\rev{0.396--1.437}) quantifies distributional movement toward the base model.

\subsubsection{Safety}

\begin{table}[h!]
\centering
\caption{\textbf{Safety} results. ``Trig'' denotes LCDD+Trigger. All behavioral metric values in \%. HB = HarmBench ASR ($\downarrow$ safer); WG~Ref.\ = WildGuard refusal rate ($\uparrow$ safer); WG~Harm.\ = WildGuard harmfulness rate ($\downarrow$ safer). Sparsity is the LCDD mask sparsity. KL values are computed on SFT-generated references; the $P\!\parallel\!Q$ column indicates the two distributions compared.}
\label{tab:safety}
\resizebox{0.9\textwidth}{!}{
\setlength{\tabcolsep}{5pt}
\begin{tabular}{@{}lr l ccccc lc@{}}
\toprule
& & & \multicolumn{3}{c}{Safety Metrics} & \multicolumn{2}{c}{Utility Metrics}
& \multicolumn{2}{c}{KL Divergence} \\
\cmidrule(lr){4-6}\cmidrule(lr){7-8}\cmidrule(lr){9-10}
Model & Sparsity & Condition
  & HB $\downarrow$ & WG Ref. $\uparrow$ & WG Harm. $\downarrow$ & MMLU & HellaSwag
  & $P\!\parallel\!Q$ & Value \\
\midrule
\multirow{4}{*}{Qwen3-0.6B}
  & \multirow{4}{*}{\rev{74.48\%}}
  & Base        & 43.0 & 32.0 & 52.0 & 40.3 & 37.5 & & \\
& & SFT    &  1.5 & 94.0 &  2.0 & 46.0 & 38.7
    & SFT$\,\|\,$LCDD        & 0.069 \\
& & LCDD        &  2.5 & 88.5 &  3.5 & 42.2 & 38.1
    & SFT$\,\|\,$Trig   & 0.120 \\
& & Trig   & \rev{32.5} & \rev{37.0} & \rev{40.5} & 45.1 & 37.9
    & Base$\,\|\,$Trig  & 0.196 \\
\midrule
\multirow{4}{*}{DeepSeek-8B}
  & \multirow{4}{*}{81.38\%}
  & Base        & 19.0 &  47.5 & 38.5 & 54.1 & 55.6 & & \\
& & SFT    &  0.0 & 100.0 &  0.0 & 55.3 & 57.1
    & SFT$\,\|\,$LCDD        & \rev{0.254} \\
& & LCDD        &  1.5 &  96.5 &  2.0 & 54.8 & 56.3
    & SFT$\,\|\,$Trig   & \rev{0.390} \\
& & Trig   & 13.0 &  56.5 & 30.5 & 49.2 & 54.1
    & Base$\,\|\,$Trig  & \rev{0.305} \\
\midrule
\multirow{4}{*}{Mistral-7B}
  & \multirow{4}{*}{59.34\%}
  & Base        & 58.5 &  25.5 & 59.5 & 59.8 & 65.8 & & \\
& & SFT    &  0.0 & 100.0 &  0.0 & 50.1 & 53.9
    & SFT$\,\|\,$LCDD        & \rev{0.266} \\
& & LCDD        &  1.0 &  97.5 &  1.0 & 57.0 & 63.6
    & SFT$\,\|\,$Trig   & \rev{0.619} \\
& & Trig   & 17.0 &  69.0 & 21.5 & 54.1 & 62.3
    & Base$\,\|\,$Trig  & 1.282 \\
\midrule
\multirow{4}{*}{Vicuna-7B}
  & \multirow{4}{*}{\rev{84.41\%}}
  & Base        & 20.5 & 76.5 & 20.5 & 48.5 & 56.6 & & \\
& & SFT    &  0.0 & 99.0 &  0.0 & 47.1 & 56.5
    & SFT$\,\|\,$LCDD        & 0.192 \\
& & LCDD        &  2.0 & 98.0 &  1.5 & 48.6 & 56.8
    & SFT$\,\|\,$Trig   & \rev{1.085} \\
& & Trig   & 22.0 & 74.5 & 23.0 & 47.9 & 56.5
    & Base$\,\|\,$Trig  & 0.014 \\
\bottomrule
\end{tabular}
}
\end{table}

As shown in Table~\ref{tab:safety}, all four models maintain high safety under LCDD (WG refusal 88.5--98.0\%; HB ASR 1.0--2.5\%), confirming that LCDD successfully compresses safety behavior into the sparse carrier. Under trigger, safety degrades along both axes for all models: refusal rates fall and attack success rates rise.

\rev{Reversion is most complete on Vicuna-7B, whose refusal rate under trigger (74.5\%) returns to the base-model level (76.5\%) with near-identical output distributions (KL(Base$\|$Trig) = 0.014). Qwen3-0.6B shows the largest behavioral suppression, with WG refusal falling by 51.5pp from 88.5\% to 37.0\%.}

Utility is stable under LCDD for all models and shows only modest degradation under trigger (at most 5.6pp on MMLU for DeepSeek), indicating that the trigger \rev{suppresses the target behavior} without broadly disrupting general language modeling.

\subsubsection{Shakespeare Style}

\begin{table}[h!]
\centering
\caption{\textbf{Shakespeare style} results. ``Trig'' denotes LCDD+Trigger.
Judge = LLM-judge score (0--5,
$\uparrow$ more Shakespearean), evaluated by Qwen3-14B
(rubric in Appendix~\ref{app:exp-details-eval}).
MMLU and HellaSwag values in \%.
KL values computed on SFT-generated references;
the $P\!\parallel\!Q$ column indicates the distributions compared.
Sparsity is the LCDD mask sparsity.}
\label{tab:shakespeare}
\resizebox{0.7\textwidth}{!}{%
\setlength{\tabcolsep}{5pt}
\begin{tabular}{@{}lr l ccc lc@{}}
\toprule
& & & Style & \multicolumn{2}{c}{Utility Metrics}
& \multicolumn{2}{c}{KL Divergence} \\
\cmidrule(lr){4-4}\cmidrule(lr){5-6}\cmidrule(lr){7-8}
Model & Sparsity & Condition
  & Judge $\uparrow$ & MMLU & HellaSwag
  & $P\!\parallel\!Q$ & Value \\
\midrule
\multirow{4}{*}{Qwen3-0.6B}
  & \multirow{4}{*}{\rev{41.20\%}}
  & Base       & 0.03 & 40.3 & 37.5 & & \\
& & SFT   & 0.79 & 45.5 & 38.0
    & SFT$\,\|\,$LCDD        & 0.265 \\
& & LCDD       & 0.44 & 41.8 & 38.6
    & SFT$\,\|\,$Trig   & 2.405 \\
& & Trig  & 0.04 & \rev{42.7} & \rev{37.6}
    & Base$\,\|\,$Trig  & 0.446 \\
\midrule
\multirow{4}{*}{DeepSeek-8B}
  & \multirow{4}{*}{54.54\%}
  & Base       & 0.01 & 54.1 & \rev{55.6} & & \\
& & SFT   & 1.73 & 54.5 & 56.1
    & SFT$\,\|\,$LCDD        & 1.520 \\
& & LCDD       & 0.98 & 55.4 & 56.8
    & SFT$\,\|\,$Trig   & 4.301 \\
& & Trig  & 0.005 & 37.4 & 51.5
    & Base$\,\|\,$Trig  & 0.531 \\
\midrule
\multirow{4}{*}{Mistral-7B}
  & \multirow{4}{*}{50.46\%}
  & Base       & 0.07 & 59.8 & 65.8 & & \\
& & SFT   & 1.75 & 47.9 & 55.0
    & SFT$\,\|\,$LCDD        & 0.853 \\
& & LCDD       & 1.18 & 55.4 & 62.7
    & SFT$\,\|\,$Trig   & 2.643 \\
& & Trig  & 0.07 & 42.3 & 53.5
    & Base$\,\|\,$Trig  & 2.052 \\
\midrule
\multirow{4}{*}{Vicuna-7B}
  & \multirow{4}{*}{29.88\%}
  & Base       & 0.05 & 48.5 & 56.6 & & \\
& & SFT   & 0.80 & 46.6 & 56.1
    & SFT$\,\|\,$LCDD        & 0.636 \\
& & LCDD       & 0.26 & 47.3 & 56.0
    & SFT$\,\|\,$Trig   & 1.702 \\
& & Trig  & 0.04 & 45.8 & 55.8
    & Base$\,\|\,$Trig  & 0.094 \\
\bottomrule
\end{tabular}%
}
\vspace{-0.8em}
\end{table}

The Shakespeare task is the most demanding for LCDD: achieved sparsity is \rev{the lowest of the three tasks on every model} (\rev{29.9--54.5\%}), and \emph{carrier fidelity} (how closely the sparse carrier reproduces full SFT behavior before intervention, measured by KL(SFT$\|$LCDD) and the task-level Judge score) is correspondingly reduced: LCDD Judge scores (\rev{0.26--1.18}) recover only part of the SFT-level style signal (0.79--1.75). This is consistent with the expectation that distributional stylistic behavior is more diffusely encoded across parameters than fixed-response or conditional refusal behaviors.

\rev{The trigger suppresses style in all four cases:} Judge@Trig falls to near-base levels (0.005--0.07) regardless of the LCDD starting point. KL(SFT$\|$Trig) is substantially larger than KL(SFT$\|$LCDD) in all cases, confirming that post-trigger behavior diverges from SFT well beyond the LCDD baseline.

\subsubsection{Cross-Task Observations}

\rev{Three patterns emerge across the 12 model--task combinations:}

\begin{itemize}[leftmargin=*]
    \item \textbf{Sparsity reflects task localizability.} \rev{On every model, the style behavior yields the lowest sparsity of the three tasks (30--55\%, versus 38--84\% for Fixed Response and safety).} \rev{This reflects the structural demands of the behavior:} distributional stylistic behavior requires broader representational support and resists aggressive compression.

    \item \textbf{\rev{Reversal is consistent across all combinations.}} \rev{In all 12 combinations, the trigger moves the model away from SFT behavior and toward the base model.}

    \item \textbf{\rev{Reversal does not require full carrier fidelity.}} In the Shakespeare task, where carrier fidelity is lower than in Fixed Response or safety runs, the trigger still \rev{reverts the style to near-base levels} in all four models.
\end{itemize}
Together, these results answer RQ1 and RQ2 affirmatively. LCDD successfully crafts sparse carriers that preserve SFT behavior under normal inference across all three behavior types and four model families. SFT-Eraser achieves reliable trigger-based reversal in all 12 model-task combinations.

\subsection{Ablation Study}\label{sec:ablation2}

\begin{revblock}
We investigate RQ3 by applying the same trigger optimization to LCDD models and to standard SFT models. We additionally verify the necessity of joint weight optimization in LCDD in Appendix~\ref{app:ablation1}.

\paragraph{Setup.}
We cross two factors: model structure (LCDD vs.\ plain SFT) and trigger objective. The circuit-targeted objective is the full SFT-Eraser loss $\mathcal{L}_{\text{trigger}} = \mathcal{L}_{\text{MSE}} + \alpha\,\mathcal{L}_{\text{KL}} + \beta\,\mathcal{L}_{2}$\rev{, and} the output-only objective removes $\mathcal{L}_{\text{MSE}}$, leaving $\alpha\,\mathcal{L}_{\text{KL}} + \beta\,\mathcal{L}_{2}$. When the circuit-targeted objective is applied to a plain SFT model, it matches the same channel set $\mathcal{C}_{\text{write}}$ extracted from the corresponding LCDD model. We run all four conditions on six model--task combinations covering all four model families and all three behavior types, and report KL(Base$\|$Trig), which measures how far the triggered model is from the base model (Section~\ref{sec:setup}). Full results, including KL(SFT$\|$Trig), are given in Appendix~\ref{app:ablation-full}.

\begin{table}[h!]
\centering
\ifshowrev\color{red}\fi
\caption{Structural ablation: KL(Base$\|$Trig) ($\downarrow$, closer to the base model) after applying the same trigger optimization to the LCDD model and to the plain SFT model. ``Circuit'' is the full SFT-Eraser objective; ``Output'' removes $\mathcal{L}_{\text{MSE}}$. Sparsity is that of the LCDD model.}
\label{tab:ab2}
\resizebox{0.85\textwidth}{!}{%
\setlength{\tabcolsep}{6pt}
\begin{tabular}{ll r cc cc}
\toprule
& & & \multicolumn{2}{c}{LCDD + trigger} & \multicolumn{2}{c}{Plain SFT + trigger} \\
\cmidrule(lr){4-5}\cmidrule(lr){6-7}
Model & Task & Sparsity & Circuit & Output & Circuit & Output \\
\midrule
DeepSeek-8B & Safety       & 81.38\% & \rev{0.305} & 0.322 & \rev{0.535} & \rev{0.507} \\
Vicuna-7B   & Safety       & 84.41\% & 0.014 & 0.014 & \rev{0.363} & \rev{0.293} \\
Qwen3-0.6B  & Safety       & 74.48\% & 0.196 & 0.182 & 0.361 & 0.330 \\
Mistral-7B  & Safety       & 59.34\% & 1.282 & \rev{1.247} & \rev{1.424} & \rev{1.304} \\
DeepSeek-8B & Fixed Resp.  & 73.38\% & \rev{1.039} & \rev{0.794} & 4.151 & \rev{2.452} \\
DeepSeek-8B & Shakespeare  & 54.54\% & 0.531 & 0.604 & 1.092 & 1.094 \\
\bottomrule
\end{tabular}
}
\end{table}

\paragraph{Finding: the crafted structure drives reversion toward the base model.}
As shown in Table~\ref{tab:ab2}, in every one of the six combinations both LCDD conditions reach a lower KL(Base$\|$Trig) than both plain-SFT conditions. The triggered LCDD model is closer to the base model regardless of which trigger objective is used, across four model families and all three behavior types. \rev{Both models encode the same SFT behavior and differ in whether its weight changes are confined to a carrier.} The crafted structure is therefore what brings triggered outputs back toward the base model.
\end{revblock}

\section{Conclusion}\label{sec:conclusion}

We investigated whether SFT-induced behaviors can be deliberately compressed into sparse behavioral carriers\rev{, so that they are} preserved under normal inference yet reversibly suppressible by an input trigger without weight modification. LCDD crafts such carriers through utility-budgeted joint optimization of routing masks and weight deltas\rev{, and} SFT-Eraser \rev{reverts the SFT-induced behavior toward the base model} via activation matching. \rev{The resulting carriers sustain the target behaviors and are, by construction, necessary for them.}
Our experiments and ablations establish that joint weight optimization is required to \rev{reach high carrier sparsity}, and that the \rev{crafted structure is what drives reversion toward the base model}.

\paragraph{Discussion.} Beyond \rev{reversal itself}, the sparse carrier framework suggests several longer-term directions. Crafted carriers provide a controlled testbed for interpretability research: unlike post-hoc circuit discovery, a carrier with known structure enables systematic study of intervention transferability and behavioral robustness. The constructability of a behavior may itself serve as a diagnostic property, since behaviors that resist LCDD compression are likely more diffusely encoded. \rev{This connects} to a minimum description length interpretation~\citep{rissanen1978modeling}, in which carrier sparsity is a computable proxy for the description complexity of the SFT behavioral transformation. \rev{Because LCDD operates on the weight delta $\Delta W = W_{\text{ft}} - W_{\text{base}}$ and does not depend on how that delta was produced, the same pipeline applies directly to checkpoints from preference-based or reinforcement-learning post-training such as DPO, PPO, or GRPO. \rev{Studying} carriers systematically across post-training methods is a natural next step.} \rev{Another direction is} how pretraining scale affects the achievable sparsity--utility frontier for crafted carriers.

\paragraph{Limitations.}
Two limitations bound the current study. First, \rev{SFT-Eraser optimizes a soft prompt in continuous embedding space, and extending reversal to discrete hard-token triggers requires further investigation.} \rev{Second, carrier fidelity depends on how compressible the target behavior is. \rev{Extending} LCDD to more complex abilities, such as multi-step reasoning, is left to future work.}

\begin{revblock}
\paragraph{Broader impact.}
The same methodology could be used to weaken safety behavior in a sparsified model. This risk arises from sparsification itself: any pipeline that compresses a fine-tuned behavior into a small parameter subset also concentrates that behavior into a small attack surface. Our results also point to a concrete mitigation. Under identical trigger optimization, plain SFT models, whose weight changes are distributed across the network, remain further from their base models than \rev{LCDD models} do (Section~\ref{sec:ablation2}), and sparser carriers revert more completely (Appendix~\ref{app:sensitivity}). Keeping the fine-tuning delta distributed, or not releasing sparsified checkpoints, therefore avoids this exposure. All experiments use publicly released base models and public datasets (WildJailbreak, HarmBench, and WildGuard), every model we trigger is one we fine-tuned ourselves, and no deployed system is involved. We release the training and analysis code needed to reproduce the method, but not optimized trigger embeddings. Applying a trigger requires white-box access to a sparsified checkpoint, and an attacker with such access already holds stronger capabilities.
\end{revblock}

\begin{ack}
\rev{Yuping Lin, Pengfei He, Yingqian Cui, and Hui Liu are supported by the National Science Foundation (NSF) under grant numbers IIS-2212032, IIS-2212144, IIS-2504089, DUE-2234015, CNS-2246050, and DRL-2405483, the Michigan Department of Agriculture and Rural Development, and the U.S. Department of Energy, Office of Science, Office of Advanced Scientific Computing Research, under Award Number DE-SC0026773. Yue Xing is supported by the National Science Foundation (NSF) under grant number DMS-2515194, the National Institutes of Health (NIH) under grant number R21AI203026, Coefficient Giving, and Tutu AI Ventures LLC.}
\end{ack}

\bibliographystyle{unsrtnat}
\bibliography{ref}

\appendix

\section{Gate--Weight Equivalence: Full Derivations}
\label{app:gate-equiv}

We prove that the activation-gating formulation used in LCDD is
equivalent to the structured weight masking in
Section~\ref{sec:method_prelim}.
Throughout, $\Delta W = W_{\text{ft}} - W_{\text{base}}$,
$M \in \{0,1\}^{\text{shape}(W)}$ is a binary mask, and
$\phi(\cdot)$ denotes the layer nonlinearity.

\subsection{FFN Layer}
\label{app:gate-ffn}

Let $\mathbf{x}\in\mathbb{R}^{d}$ be the layer input,
$W_{\text{up}}^{\text{base}}\in\mathbb{R}^{d\times d_{\text{ffn}}}$
and
$W_{\text{down}}^{\text{base}}\in\mathbb{R}^{d_{\text{ffn}}\times d}$
the base FFN weights, and $\Delta W_{\text{up}},\Delta W_{\text{down}}$
the corresponding fine-tuning increments.
Three binary gate vectors control the delta path:

\medskip
\begin{center}
\begin{tabular}{ll}
$\mathbf{m}_{\text{read}}\in\{0,1\}^{d}$ &
  gates input dimensions (residual read) \\
$\mathbf{m}_{\text{hidden}}\in\{0,1\}^{d_{\text{ffn}}}$ &
  gates hidden (intermediate) dimensions \\
$\mathbf{m}_{\text{write}}\in\{0,1\}^{d}$ &
  gates output dimensions (residual write)
\end{tabular}
\end{center}

\paragraph{Activation-gating forward pass.}
\begin{align}
\mathbf{h} &= \phi\!\left(
    \mathbf{x}\,W_{\text{up}}^{\text{base}}
  + \bigl(\mathbf{x}\odot\mathbf{m}_{\text{read}}\bigr)
    \Delta W_{\text{up}}\odot\mathbf{m}_{\text{hidden}}
  \right) \label{eq:ffn-h} \\
\Delta\mathbf{y} &=
  \bigl(\mathbf{h}\odot\mathbf{m}_{\text{hidden}}\bigr)
  \Delta W_{\text{down}}\odot\mathbf{m}_{\text{write}}
  \label{eq:ffn-dy}
\end{align}

\paragraph{Equivalent weight masks.}
\begin{equation}
  M_{\text{up}}[i,j]   = m_{\text{read}}[i]  \cdot m_{\text{hidden}}[j],
  \qquad
  M_{\text{down}}[i,j] = m_{\text{hidden}}[i] \cdot m_{\text{write}}[j].
  \label{eq:ffn-masks}
\end{equation}

\begin{proof}[Proof for $W_{\text{up}}$]
The $j$-th pre-activation under activation gating is
\begin{align}
\mathrm{pre}_{j}
&= \sum_{i} x_i\,W_{\text{up}}^{\text{base}}[i,j]
 + \Bigl(\sum_{i}(x_i\cdot m_{\text{read}}[i])\,
         \Delta W_{\text{up}}[i,j]\Bigr)\,m_{\text{hidden}}[j] \notag\\
&= \sum_{i} x_i\,W_{\text{up}}^{\text{base}}[i,j]
 + \sum_{i} x_i\cdot m_{\text{read}}[i]\cdot m_{\text{hidden}}[j]
   \cdot\Delta W_{\text{up}}[i,j].
\end{align}
Under weight masking with $M_{\text{up}}[i,j]=m_{\text{read}}[i]\cdot m_{\text{hidden}}[j]$:
\begin{align}
\mathrm{pre}_{j}
&= \sum_{i} x_i\Bigl(W_{\text{up}}^{\text{base}}[i,j]
  + M_{\text{up}}[i,j]\cdot\Delta W_{\text{up}}[i,j]\Bigr) \notag\\
&= \sum_{i} x_i\,W_{\text{up}}^{\text{base}}[i,j]
  + \sum_{i} x_i\cdot m_{\text{read}}[i]\cdot m_{\text{hidden}}[j]
    \cdot\Delta W_{\text{up}}[i,j].
\quad\checkmark
\end{align}
\end{proof}

\begin{proof}[Proof for $W_{\text{down}}$]
The $j$-th output delta under activation gating is
\begin{equation}
\Delta y_j
= \Bigl(\sum_{i} h_i\cdot m_{\text{hidden}}[i]
        \cdot\Delta W_{\text{down}}[i,j]\Bigr)m_{\text{write}}[j]
= \sum_{i} h_i\cdot m_{\text{hidden}}[i]\cdot m_{\text{write}}[j]
  \cdot\Delta W_{\text{down}}[i,j].
\end{equation}
Under weight masking with $M_{\text{down}}[i,j]=m_{\text{hidden}}[i]\cdot m_{\text{write}}[j]$:
\begin{equation}
\Delta y_j
= \sum_{i} h_i\cdot M_{\text{down}}[i,j]\cdot\Delta W_{\text{down}}[i,j]
= \sum_{i} h_i\cdot m_{\text{hidden}}[i]\cdot m_{\text{write}}[j]
  \cdot\Delta W_{\text{down}}[i,j].
\quad\checkmark\qquad\square
\end{equation}
\end{proof}

\subsection{Attention Layer}
\label{app:gate-attn}

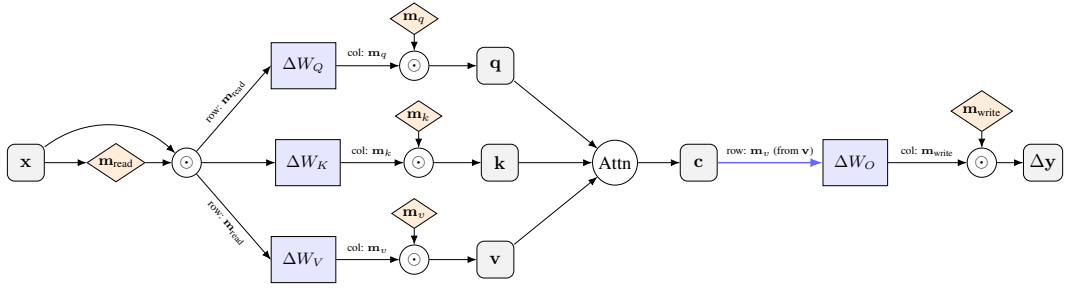
\begin{figure}[t]
\centering
\resizebox{\textwidth}{!}{%
\begin{tikzpicture}[
    node distance=1.2cm and 1.5cm,
    >=Latex,
    tensor/.style={draw, rectangle, rounded corners,
                   fill=gray!10, minimum height=2em,
                   minimum width=2em, align=center},
    weight/.style={draw, rectangle, fill=blue!10,
                   minimum height=2.5em, minimum width=3.5em,
                   align=center, font=\small\bfseries},
    gate/.style={draw, diamond, fill=orange!15, aspect=1.5,
                 inner sep=1pt, font=\footnotesize},
    op/.style={draw, circle, inner sep=2pt, fill=white},
]

\node[tensor] (x)       {$\mathbf{x}$};
\node[gate,   right=0.8cm of x]        (g_read)    {$\mathbf{m}_{\text{read}}$};
\node[op,     right=0.5cm of g_read]   (mul_read)  {$\odot$};
\draw[->] (x)       -- (g_read);
\draw[->] (g_read)  -- (mul_read);
\draw[->] (x) to[bend left=40] (mul_read);

\node[weight, above right=1.2cm and 1.4cm of mul_read] (WQ) {$\Delta W_Q$};
\node[weight, right=1.4cm              of mul_read]    (WK) {$\Delta W_K$};
\node[weight, below right=1.2cm and 1.4cm of mul_read] (WV) {$\Delta W_V$};

\draw[->] (mul_read) -- (WQ.west)
    node[midway, above, sloped, font=\tiny] {row: $\mathbf{m}_{\text{read}}$};
\draw[->] (mul_read) -- (WK.west);
\draw[->] (mul_read) -- (WV.west)
    node[midway, below, sloped, font=\tiny] {row: $\mathbf{m}_{\text{read}}$};

\node[op,   right=1.2cm of WQ] (mul_q) {$\odot$};
\node[gate, above=0.3cm of mul_q]  (g_q)  {$\mathbf{m}_q$};
\node[op,   right=1.2cm of WK] (mul_k) {$\odot$};
\node[gate, above=0.3cm of mul_k]  (g_k)  {$\mathbf{m}_k$};
\node[op,   right=1.2cm of WV] (mul_v) {$\odot$};
\node[gate, above=0.3cm of mul_v]  (g_v)  {$\mathbf{m}_v$};

\draw[->] (WQ) -- (mul_q) node[midway, above, font=\tiny] {col: $\mathbf{m}_q$};
\draw[->] (g_q) -- (mul_q);
\draw[->] (WK) -- (mul_k) node[midway, above, font=\tiny] {col: $\mathbf{m}_k$};
\draw[->] (g_k) -- (mul_k);
\draw[->] (WV) -- (mul_v) node[midway, above, font=\tiny] {col: $\mathbf{m}_v$};
\draw[->] (g_v) -- (mul_v);

\node[tensor, right=0.9cm of mul_q] (Q) {$\mathbf{q}$};
\node[tensor, right=0.9cm of mul_k] (K) {$\mathbf{k}$};
\node[tensor, right=0.9cm of mul_v] (V) {$\mathbf{v}$};
\draw[->] (mul_q) -- (Q);
\draw[->] (mul_k) -- (K);
\draw[->] (mul_v) -- (V);

\node[op, right=1.4cm of K] (attn) {Attn};
\draw[->] (Q) -- (attn);
\draw[->] (K) -- (attn);
\draw[->] (V) -- (attn);

\node[tensor, right=0.8cm of attn] (ctx) {$\mathbf{c}$};
\draw[->] (attn) -- (ctx);

\node[weight, right=2cm of ctx] (WO) {$\Delta W_O$};
\draw[->, thick, blue!60] (ctx) -- (WO)
    node[midway, above, font=\tiny, black]
        {row: $\mathbf{m}_v$ (from $\mathbf{v}$)};

\node[op,     right=1.5cm of WO]     (mul_w) {$\odot$};
\node[gate,   above=0.3cm of mul_w]  (g_w)   {$\mathbf{m}_{\text{write}}$};
\node[tensor, right=0.5cm of mul_w]  (dy)    {$\Delta\mathbf{y}$};
\draw[->] (WO)  -- (mul_w)
    node[midway, above, font=\tiny] {col: $\mathbf{m}_{\text{write}}$};
\draw[->] (g_w) -- (mul_w);
\draw[->] (mul_w) -- (dy);

\end{tikzpicture}%
}
\caption{Delta-path computation graph for the attention layer with
activation gating.
Input $\mathbf{x}$ is gated by $\mathbf{m}_{\text{read}}$ before
entering all projection deltas; Q/K/V outputs are independently gated
by $\mathbf{m}_q,\mathbf{m}_k,\mathbf{m}_v$.
The row mask of $\Delta W_O$ is $\mathbf{m}_v$ (not
$\mathbf{m}_{\text{read}}$) because $W_O$ reads from the context
vector $\mathbf{c}$, which is derived from value projections.}
\label{fig:attn-gate}
\end{figure}

Let $\mathbf{x}\in\mathbb{R}^{d}$ be the layer input,
$W_Q^{\text{base}},W_K^{\text{base}},W_V^{\text{base}}
 \in\mathbb{R}^{d\times d_{\text{inner}}}$
the base projection weights, and
$W_O^{\text{base}}\in\mathbb{R}^{d_{\text{inner}}\times d}$
the output projection weight.
Five binary gate vectors control the delta path
(Figure~\ref{fig:attn-gate}):

\medskip
\begin{center}
\begin{tabular}{ll}
$\mathbf{m}_{\text{read}}\in\{0,1\}^{d}$ &
  gates input dimensions \\
$\mathbf{m}_q,\mathbf{m}_k,\mathbf{m}_v
 \in\{0,1\}^{d_{\text{inner}}}$ &
  gate Q/K/V head dimensions independently \\
$\mathbf{m}_{\text{write}}\in\{0,1\}^{d}$ &
  gates output dimensions
\end{tabular}
\end{center}

\paragraph{Activation-gating forward pass.}
\begin{align}
\mathbf{q} &= \mathbf{x}\,W_Q^{\text{base}}
  + (\mathbf{x}\odot\mathbf{m}_{\text{read}})\,\Delta W_Q
    \odot\mathbf{m}_q \label{eq:attn-q}\\
\mathbf{k} &= \mathbf{x}\,W_K^{\text{base}}
  + (\mathbf{x}\odot\mathbf{m}_{\text{read}})\,\Delta W_K
    \odot\mathbf{m}_k \label{eq:attn-k}\\
\mathbf{v} &= \mathbf{x}\,W_V^{\text{base}}
  + (\mathbf{x}\odot\mathbf{m}_{\text{read}})\,\Delta W_V
    \odot\mathbf{m}_v \label{eq:attn-v}\\
\mathbf{c} &= \mathrm{Attention}(\mathbf{q},\mathbf{k},\mathbf{v})
  \label{eq:attn-c}\\
\Delta\mathbf{y} &=
  (\mathbf{c}\odot\mathbf{m}_v)\,\Delta W_O\odot\mathbf{m}_{\text{write}}
  \label{eq:attn-dy}
\end{align}

The row gate for $\Delta W_O$ in Eq.~\eqref{eq:attn-dy} is
$\mathbf{m}_v$, not $\mathbf{m}_{\text{read}}$.
This follows from data flow: $W_O$ reads from the context vector
$\mathbf{c}$, which is assembled from value projections.
Zeroing value channel $i$ (i.e.\ $m_v[i]=0$) should suppress row $i$
of $\Delta W_O$; using $\mathbf{m}_{\text{read}}$ here would be
structurally inconsistent.

\paragraph{Equivalent weight masks.}
\begin{equation}
\begin{aligned}
  M_Q[i,j] &= m_{\text{read}}[i]\cdot m_q[j],\quad
  M_K[i,j] = m_{\text{read}}[i]\cdot m_k[j],\\
  M_V[i,j] &= m_{\text{read}}[i]\cdot m_v[j],\quad
  M_O[i,j] = m_v[i]\cdot m_{\text{write}}[j].
\end{aligned}
  \label{eq:attn-masks}
\end{equation}

\begin{proof}[Proof for $W_Q$, $W_K$, $W_V$]
The arguments are identical; we take $W_Q$ as representative.
The $j$-th component of the delta query under activation gating:
\begin{equation}
\Delta q_j
= \Bigl(\sum_{i}(x_i\cdot m_{\text{read}}[i])
        \cdot\Delta W_Q[i,j]\Bigr)m_q[j]
= \sum_{i} x_i\cdot m_{\text{read}}[i]\cdot m_q[j]
  \cdot\Delta W_Q[i,j].
\end{equation}
Under weight masking with $M_Q[i,j]=m_{\text{read}}[i]\cdot m_q[j]$:
\begin{equation}
\Delta q_j
= \sum_{i} x_i\cdot M_Q[i,j]\cdot\Delta W_Q[i,j]
= \sum_{i} x_i\cdot m_{\text{read}}[i]\cdot m_q[j]
  \cdot\Delta W_Q[i,j].\quad\checkmark
\end{equation}
\end{proof}

\begin{proof}[Proof for $W_O$]
Let $\mathbf{c}\in\mathbb{R}^{d_{\text{inner}}}$ be the context
vector.
The $j$-th output delta under activation gating:
\begin{equation}
\Delta y_j
= \Bigl(\sum_{k} c_k\cdot m_v[k]
        \cdot\Delta W_O[k,j]\Bigr)m_{\text{write}}[j]
= \sum_{k} c_k\cdot m_v[k]\cdot m_{\text{write}}[j]
  \cdot\Delta W_O[k,j].
\end{equation}
Under weight masking with $M_O[k,j]=m_v[k]\cdot m_{\text{write}}[j]$:
\begin{equation}
\Delta y_j
= \sum_{k} c_k\cdot M_O[k,j]\cdot\Delta W_O[k,j]
= \sum_{k} c_k\cdot m_v[k]\cdot m_{\text{write}}[j]
  \cdot\Delta W_O[k,j].\quad\checkmark\qquad\square
\end{equation}
\end{proof}

\subsection{Summary}
\label{app:gate-summary}

Table~\ref{tab:gate-map} collects the row and column gate assignments
for all six delta weight matrices per transformer layer.
Two structural observations follow directly from the assignment.
First, $\mathbf{m}_{\text{read}}$ and $\mathbf{m}_{\text{write}}$ act
as global input/output interfaces: any gate-group optimisation that
suppresses $\mathbf{m}_{\text{read}}$ simultaneously prunes the
row dimension of all four attention projection deltas and the row of
$\Delta W_{\text{up}}$.
Second, $\mathbf{m}_v$ is shared between $W_V$ (column) and $W_O$
(row), enforcing structural consistency: a value channel that is
pruned in the projection stage is also pruned in the readout stage,
preventing information leakage through the output projection.

\begin{table}[h]
\centering
\caption{Gate-to-weight-mask assignment per transformer layer.
The 8 gate vectors total 3 (FFN) $+$ 5 (Attention).
$d$ = residual-stream dimension;
$d_{\text{ffn}}$ = FFN hidden dimension;
$d_{\text{inner}}$ = attention head dimension.}
\label{tab:gate-map}
\setlength{\tabcolsep}{8pt}
\begin{tabular}{llccc}
\toprule
\textbf{Module} & \textbf{Weight matrix} & \textbf{Shape} &
\textbf{Row gate} & \textbf{Col gate} \\
\midrule
\multirow{2}{*}{FFN}
  & $W_{\text{up}}$   & $d\times d_{\text{ffn}}$
    & $\mathbf{m}_{\text{read}}$   & $\mathbf{m}_{\text{hidden}}$ \\
  & $W_{\text{down}}$ & $d_{\text{ffn}}\times d$
    & $\mathbf{m}_{\text{hidden}}$ & $\mathbf{m}_{\text{write}}$  \\
\midrule
\multirow{4}{*}{Attention}
  & $W_Q$ & $d\times d_{\text{inner}}$
    & $\mathbf{m}_{\text{read}}$ & $\mathbf{m}_q$ \\
  & $W_K$ & $d\times d_{\text{inner}}$
    & $\mathbf{m}_{\text{read}}$ & $\mathbf{m}_k$ \\
  & $W_V$ & $d\times d_{\text{inner}}$
    & $\mathbf{m}_{\text{read}}$ & $\mathbf{m}_v$ \\
  & $W_O$ & $d_{\text{inner}}\times d$
    & $\mathbf{m}_v$             & $\mathbf{m}_{\text{write}}$ \\
\bottomrule
\end{tabular}
\end{table}

\section{Gate Learning: Heaviside Binarization and STE}
\label{app:gate-learning}

Each gate logit $\theta_i \in \mathbb{R}$ is binarized during the forward pass via the Heaviside function:
\begin{equation}
  m_i = \mathbf{1}[\theta_i > 0] \;\in\; \{0,1\}.
\end{equation}
Since $\mathbf{1}[\cdot]$ has zero gradient almost everywhere, backpropagation is infeasible directly. We apply the straight-through estimator (STE)~\citep{bengio2013estimating}: during the backward pass, the gradient with respect to $\theta_i$ is approximated as
\begin{equation}
  \frac{\partial \mathcal{L}}{\partial \theta_i}
  \;\approx\;
  \frac{\partial \mathcal{L}}{\partial m_i}
  \cdot \sigma'(\theta_i),
  \qquad
  \sigma'(\theta_i) = \sigma(\theta_i)\bigl(1-\sigma(\theta_i)\bigr),
\end{equation}
while the forward pass retains the exact binary value. We use Heaviside + STE rather than HardConcrete-based $L_0$ relaxation~\citep{louizos2017learning}, which introduces a training--inference inconsistency via continuous masks; \citep{gao2025weight} report that Heaviside-based approaches consistently outperform the HardConcrete variant in sparse training settings.

\section{LCDD Controller: Full Algorithm}
\label{app:lcdd-algo}

We give here the complete pseudocode for the LCDD adaptive dual-inspired controller described in Section~\ref{sec:method_lcdd}. The algorithm instantiates the adaptive penalty method for the constrained objective (Equation~\ref{eq:lcdd-obj}), combining a linear sparsity warmup, an exponential moving average (EMA)-based budget threshold, and a multiplicative multiplier update driven by the normalized constraint violation ratio.

\begin{algorithm}[h]
\caption{LCDD: Utility-Budgeted Sparse Carrier Crafting}
\label{alg:lcdd}
\begin{algorithmic}[1]
\Require Base model $W_{\text{base}}$, fine-tuned model $W_{\text{ft}}$,
         budget ratio $r$, warmup steps $T_w$, multiplier bounds
         $\lambda_{\min}, \lambda_{\max}$, multiplier step size $\eta_\lambda$,
         EMA decay $\beta$
\Ensure  Sparse mask $M^*$, adapted weight increment $\Delta W^*$

\State Initialize $\Delta W \gets W_{\text{ft}} - W_{\text{base}}$,
       gate logits $\theta_i \gets 0\ \forall i$,
       multiplier $\lambda_0$,
       EMA estimate $\widehat{\mathcal{L}}_0 \gets \mathcal{L}_{\text{task}}$

\For{each training step $t = 1, 2, \ldots$}

    \State \textbf{// Sparsity warmup}
    \State $\rho_t \gets \min(1,\ t / T_w)$

    \State \textbf{// Forward pass with binary gates}
    \State $m_i \gets \mathbf{1}[\theta_i > 0]\ \forall i$
    \State Compute $\mathcal{L}_{\text{task}}$ and
           $\mathcal{L}_{\text{sparsity}} = \sum_i \sigma(\theta_i)$

    \State \textbf{// Combined loss}
    \State $\mathcal{L} \gets \mathcal{L}_{\text{task}}
           + \lambda_t\,\rho_t\,\mathcal{L}_{\text{sparsity}}$

    \State \textbf{// Update $\theta$ via STE, update $\Delta W$ via standard grad}
    \State $\theta,\, \Delta W \gets \textsc{BackwardStep}(\mathcal{L})$

    \State \textbf{// Update EMA of task loss}
    \State $\widehat{\mathcal{L}}_t \gets
           \beta\,\widehat{\mathcal{L}}_{t-1}
           + (1-\beta)\,\mathcal{L}_{\text{task}}$

    \If{$t = T_w$}  \Comment{Set budget after warmup stabilizes}
        \State $\epsilon \gets \widehat{\mathcal{L}}_{T_w}
               \cdot (1 + r)$
    \EndIf

    \State \textbf{// Update multiplier via normalized constraint violation ratio}
    \State $v_t \gets
           (\widehat{\mathcal{L}}_t - \epsilon)\ /\ \epsilon$
    \State $\lambda_{t+1} \gets \mathrm{clip}\!\left(
           \lambda_t \exp(-\eta_\lambda\,v_t),\;
           \lambda_{\min},\,\lambda_{\max}\right)$

    \If{early-stop criterion met}  \Comment{Sparsity stall or budget breach}
        \State \textbf{break}
    \EndIf

\EndFor

\State \Return $M^* = \{m_i\}$,\quad $\Delta W^*$
\end{algorithmic}
\end{algorithm}

\section{Trigger Optimization Details}
\label{app:trigger-optim}

After each gradient update on $\mathcal{L}_{\text{trigger}}$, we apply PGD-style $\ell_2$ projection to keep each trigger token within a bounded embedding region:
\begin{equation}
t_i \leftarrow
\Pi_{\|\cdot\|_2 \le R}\!\left(
t_i - \eta_t \nabla_{t_i}\mathcal{L}_{\text{trigger}}
\right).
\end{equation}
This prevents the optimizer from exploiting unbounded embedding directions to satisfy the activation matching objective trivially. 
Full trigger hyperparameters are listed in Appendix~\ref{app:exp-details-trigger}.

\section{Ablation Study: Necessity of Joint Weight Optimization}
\label{app:ablation1}

We test whether jointly optimizing masks and weight deltas is strictly necessary. As an alternative, we fix $W_{\text{ft}}$ and optimize only the gate parameters $\theta$. This follows the fixed-weight masking approach used in mask-based model adaptation~\citep{mallya2018piggyback} and safety circuit attribution~\citep{yu2026safeseek}, applied here to the SFT delta path. We evaluate whether this simpler setup achieves comparable sparsity\rev{, and whether the resulting carriers remain reversible}.

\paragraph{Setup.}
We construct two mask-only variants under the same LCDD controller, budget ratio (0.30), and early-stop criterion as the main experiment. Two loss variants are tested: (1) CE-driven mask-only and (2) KL-distillation mask-only. The same trigger optimization is then applied to each checkpoint. Both variants are evaluated on DeepSeek-R1-Distill-Llama-8B under the safety task.

\begin{table}[h!]
\centering
\caption{Ablation: sparsity and trigger-based reversal on DeepSeek-R1-Distill-Llama-8B (safety task). KL metrics computed on SFT-generated references. Base WG refusal: 47.5\%; SFT WG refusal: 100.0\%.}
\label{tab:ab1}
\setlength{\tabcolsep}{5pt}
\resizebox{1\textwidth}{!}{
\begin{tabular}{l c c cc cccc}
\toprule
& & & \multicolumn{2}{c}{WG Refusal}
& \multicolumn{3}{c}{KL Divergence} \\
\cmidrule(lr){4-5}\cmidrule(lr){6-8}
Method & Stop Epoch & Sparsity
& LCDD & +Trig
& SFT$\|$LCDD & SFT$\|$Trig & Base$\|$Trig \\
\midrule
LCDD joint (main) & ---  & 81.38\%
  & 96.5\% & 56.5\% & \rev{0.254} & \rev{0.390} & \rev{0.305} \\
Mask-only (CE)    & 3.70 & 28.40\%
  & 99.0\% & 64.0\% & 0.121 & 0.262 & 0.254 \\
Mask-only (KL)    & 3.83 & 33.71\%
  & 98.0\% & 60.0\% & 0.169 & \rev{0.318} & \rev{0.196} \\
\bottomrule
\end{tabular}
}
\end{table}

\paragraph{Finding 1: mask-only cannot achieve comparable sparsity.} 
Both mask-only variants trigger early stopping by epoch $\approx$3.8. They reach only 28--34\% sparsity. This is \rev{2.4--2.9$\times$ lower than the 81.38\%} achieved by joint optimization. Without weight adaptation, mask deletions directly degrade safety behavior. The optimizer hits a hard constraint and stops within the first four epochs. Joint optimization avoids this. Weight deltas reorganize around pruned connections\rev{, keeping behavior within the utility budget}.

\begin{revblock}
\paragraph{Finding 2: joint optimization compresses the carrier while keeping it reversible.}
All three carriers remain reversible under the same trigger optimization: WG refusal falls from 96.5\% to 56.5\% for the joint carrier, and from 99.0\% to 64.0\% (CE) and from 98.0\% to 60.0\% (KL) for the mask-only carriers. Joint optimization therefore yields a carrier 2.4--2.9$\times$ sparser than masking alone while retaining reversal of comparable magnitude.

\paragraph{Synthesis.}
Joint weight optimization is necessary to reach the sparsity regime of LCDD. Masking the fixed SFT delta stalls at 28--34\% sparsity, whereas jointly adapting the weights lets the delta reorganize around pruned connections and reach 81.38\% sparsity, with the behavior kept within the utility budget and the resulting carrier reversible.
\end{revblock}

\begin{revblock}
\section{\texorpdfstring{\rev{Structural Ablation: Full Results}}{Structural Ablation: Full Results}}
\label{app:ablation-full}

Table~\ref{tab:ab2-full} extends Table~\ref{tab:ab2} with KL(SFT$\|$Trig), which measures how far the trigger moves each model away from its own fine-tuned behavior. In all six combinations and under both trigger objectives, the trigger moves the LCDD model further from SFT behavior than the plain SFT model, and brings it closer to the base model. When the circuit-targeted objective is applied to a plain SFT model, it matches the channel set $\mathcal{C}_{\text{write}}$ extracted from the LCDD model of the same model--task combination. KL values are computed with SFT-generated references as in Appendix~\ref{app:exp-details-eval}.

\begin{table}[h!]
\centering
\ifshowrev\color{red}\fi
\caption{Full structural ablation results. ``Circuit'' is the full SFT-Eraser objective; ``Output'' removes $\mathcal{L}_{\text{MSE}}$.}
\label{tab:ab2-full}
\resizebox{\textwidth}{!}{%
\setlength{\tabcolsep}{5pt}
\begin{tabular}{ll cccc cccc}
\toprule
& & \multicolumn{4}{c}{KL(Base$\|$Trig) $\downarrow$} & \multicolumn{4}{c}{KL(SFT$\|$Trig) $\uparrow$} \\
\cmidrule(lr){3-6}\cmidrule(lr){7-10}
& & \multicolumn{2}{c}{LCDD} & \multicolumn{2}{c}{Plain SFT} & \multicolumn{2}{c}{LCDD} & \multicolumn{2}{c}{Plain SFT} \\
\cmidrule(lr){3-4}\cmidrule(lr){5-6}\cmidrule(lr){7-8}\cmidrule(lr){9-10}
Model & Task & Circuit & Output & Circuit & Output & Circuit & Output & Circuit & Output \\
\midrule
DeepSeek-8B & Safety      & 0.305 & 0.322 & 0.535 & 0.507 & 0.390 & 0.438 & 0.222 & 0.209 \\
Vicuna-7B   & Safety      & 0.014 & 0.014 & 0.363 & 0.293 & 1.085 & 1.062 & 0.403 & 0.409 \\
Qwen3-0.6B  & Safety      & 0.196 & 0.182 & 0.361 & 0.330 & 0.120 & 0.133 & 0.078 & 0.070 \\
Mistral-7B  & Safety      & 1.282 & 1.247 & 1.424 & 1.304 & 0.619 & 0.619 & 0.416 & 0.467 \\
DeepSeek-8B & Fixed Resp. & 1.039 & 0.794 & 4.151 & 2.452 & 1.810 & 2.116 & 0.736 & 1.017 \\
DeepSeek-8B & Shakespeare & 0.531 & 0.604 & 1.092 & 1.094 & 4.301 & 4.171 & 3.486 & 3.294 \\
\bottomrule
\end{tabular}
}
\end{table}

\section{\texorpdfstring{\rev{Sensitivity Analysis}}{Sensitivity Analysis}}
\label{app:sensitivity}

We vary one setting at a time on Qwen3-0.6B (safety), holding all others at the values of the main experiment. KL is computed as in Appendix~\ref{app:exp-details-eval}.

\paragraph{Utility budget.}
The budget ratio $r$ sets $\epsilon = \widehat{\mathcal{L}}\,(1+r)$ in Algorithm~\ref{alg:lcdd}, where $\widehat{\mathcal{L}}$ is the EMA of the task loss at the end of warmup. As shown in Table~\ref{tab:sens-budget}, a larger budget yields a sparser carrier and a more complete reversion: from $r=0.0005$ to $r=0.10$, sparsity rises from 72.03\% to 90.61\% and KL(Base$\|$Trig) falls from 0.195 to 0.079, while KL(SFT$\|$LCDD) rises from 0.063 to 0.212. The budget is thus a single interpretable control that trades carrier fidelity for completeness of reversion.

\paragraph{Trigger length and channel selection.}
Reversion is stable in both settings (Table~\ref{tab:sens-trigger}). Across an eight-fold change in trigger length (5 to 40 tokens), KL(Base$\|$Trig) stays within 0.193--0.198. Matching all residual channels, only the FFN write channels, or only the attention write channels gives KL(Base$\|$Trig) within 0.192--0.203, matching the default residual write channels (0.196).

\begin{table}[h!]
\centering
\ifshowrev\color{red}\fi
\caption{Utility budget sweep on Qwen3-0.6B (safety). $r=0.01$ is the main-experiment setting.}
\label{tab:sens-budget}
\small
\begin{tabular}{c c c c}
\toprule
Budget ratio $r$ & Sparsity & KL(SFT$\|$LCDD) $\downarrow$ & KL(Base$\|$Trig) $\downarrow$ \\
\midrule
0.0005 & 72.03\% & 0.063 & 0.195 \\
0.003  & 72.72\% & 0.063 & 0.201 \\
0.01   & 74.48\% & 0.069 & 0.196 \\
0.04   & 81.74\% & 0.107 & 0.151 \\
0.10   & 90.61\% & 0.212 & 0.079 \\
\bottomrule
\end{tabular}
\end{table}

\begin{table}[h!]
\centering
\ifshowrev\color{red}\fi
\caption{Trigger length and matched channel set on Qwen3-0.6B (safety), using the main-experiment carrier (74.48\% sparsity). Defaults are length 20 and residual write channels.}
\label{tab:sens-trigger}
\small
\begin{tabular}{l c c}
\toprule
Setting & KL(SFT$\|$Trig) & KL(Base$\|$Trig) $\downarrow$ \\
\midrule
Length 5   & 0.136 & 0.193 \\
Length 10  & 0.131 & 0.198 \\
Length 20 (default) & 0.120 & 0.196 \\
Length 40  & 0.130 & 0.196 \\
\midrule
Residual write (default) & 0.120 & 0.196 \\
All residual (read + write) & 0.135 & 0.192 \\
FFN write only & 0.120 & 0.203 \\
Attention write only & 0.120 & 0.203 \\
\bottomrule
\end{tabular}
\end{table}
\end{revblock}

\section{Experimental Details}
\label{app:exp-details}

\subsection{Dataset Details}
\label{app:exp-details-dataset}

\paragraph{Fixed Response task (training and evaluation prompts).}
For Fixed Response training, we use the Alpaca
dataset~\citep{taori2023alpaca} and format each sample as a chat
prompt with a fixed target completion ``I don't know.''
Main runs use the first 5,000 examples from the Alpaca
\texttt{train} split (without shuffling); samples with a non-empty
secondary input field have that field appended to the instruction.
Evaluation prompts are drawn from a held-out pool by
excluding those first 5,000 examples, shuffling the remainder with
a fixed seed, and retaining only instruction-only rows
(empty secondary input field).

\paragraph{Safety task.}
We use the WildJailbreak dataset~\citep{jiang2024wildteaming} and separate harmful and
benign subsets. Main safety training uses a 1:1 harmful/benign
mixture (5,000 total), matching the intended alignment objective of
retaining utility while learning refusal behavior. Safety evaluation
uses HarmBench standard behaviors and reports HarmBench ASR plus
WildGuard refusal and harmfulness rates.

\paragraph{Shakespeare task.}
We use the Shakespearean and Modern English Conversational
Dataset~\citep{roudranil2023shakespeare}, treating the modern English
translation as input and the original Shakespearean text as target.
\rev{The dataset provides}
paired modern~/ Shakespearean dialogue turns at the utterance level.
Main runs use 5,000 training samples.

\subsection{Training Hyperparameters}
\label{app:exp-details-train}

\paragraph{Shared settings.}
All LCDD runs follow a two-phase pipeline: the SFT checkpoint
provides the fine-tuned weights $W_{\text{ft}}$, and the original
pretrained model provides the base weights $W_{\text{base}}$.
Mask parameters are optimized with SGD (momentum 0.0) at learning
rate 0.1. The SFT stage uses learning rate $2\times10^{-5}$ across
all main runs. Training sample count is 5,000 per task.

Table~\ref{tab:app-main-lcdd-hparams} lists the LCDD hyperparameters
for the 12 primary model-task runs. The column
$\eta_{\lambda,\uparrow}$ denotes a separate growth learning rate
used in asymmetric dual controller variants; ``--'' indicates the
standard symmetric update is used instead.

\begin{table}[h]
\centering
\small
\caption{LCDD hyperparameters for the 12 primary model-task runs.
$\eta_\lambda$: multiplier update step size.
$\eta_{\lambda,\uparrow}$: asymmetric growth rate
(``--'' = symmetric update).}
\label{tab:app-main-lcdd-hparams}
\resizebox{1.0\textwidth}{!}{
\begin{tabular}{lcccccc}
\toprule
Setting & budget ratio & $\lambda_0$ & $\eta_\lambda$
        & $\eta_{\lambda,\uparrow}$ & warmup steps & epochs \\
\midrule
Fixed Response (Qwen, Mistral, DeepSeek)
        & 0.30 & 1.0  & 0.1 & --    & 300 & 10 \\
Fixed Response (Vicuna)
        & 0.30 & 0.01 & 0.1 & 0.01  & 700 & 20 \\
Safety (Mistral, DeepSeek)
        & 0.30 & 0.01 & 0.1 & --    & 300 & 20 \\
Safety (Qwen)
        & 0.01 & 0.01 & 0.1 & 0.01  & 300 & 20 \\
Safety (Vicuna)
        & 0.50 & 0.01 & 1.0 & 0.005 & 300 & 20 \\
Shakespeare (Qwen, Mistral, Vicuna)
        & 0.01 & 0.01 & 0.1 & 0.01  & 700 & 20 \\
Shakespeare (DeepSeek)
        & 0.01 & 0.01 & 0.1 & 0.01  & 300 & 20 \\
\bottomrule
\end{tabular}
}
\end{table}

\paragraph{Additional stabilizers.}
For runs with asymmetric dual control, $\lambda_{\max}\in\{0.02,
0.05\}$. For Qwen and Vicuna safety and Shakespeare runs, we apply
EMA smoothing with $\beta_{\text{ema}}=0.90$ to the task loss
estimate. 

\paragraph{Compute resources.}
Each experiment run (one model-task combination) is conducted on a single NVIDIA H200 GPU (80GB).

\subsection{Trigger Optimization Hyperparameters}
\label{app:exp-details-trigger}

\paragraph{Default trigger settings.}
Trigger optimization targets residual write channels and uses the
following settings uniformly across all runs: trigger length 20,
batch size 16, learning rate 0.003, KL weight $\alpha=0.7$,
tail-$k$ KL with $k=8$, $\ell_2$ regularization weight $\beta=0.1$,
$\ell_2$ max-norm $R=1.0$, 200 prompt pairs, 2000 optimization steps.
Prompt sources are task-matched: Alpaca prompts for Fixed Response,
WildJailbreak 50/50 mix for Safety, and Shakespeare dataset prompts
for style.

\subsection{Evaluation Protocol}
\label{app:exp-details-eval}

\paragraph{Fixed-response rate.}
Evaluation uses keyword matching against the set
\{``i don't know'', ``i do not know'',
``i'm not sure'', ``i am not sure''\},
with response length capped at 40 words to separate
clean fixed responses from hedged long-form answers.

\paragraph{Shakespeare LLM-judge rubric.}
The judge prompt asks for a single integer 0--5 and is
reproduced verbatim below.

\begin{small}
\begin{verbatim}
Rate the following response on a scale of 0 to 5 for how
Shakespearean it sounds.

0 = completely modern English, no archaic features
1 = occasional archaic word but mostly modern
2 = noticeably archaic vocabulary, some Shakespearean phrasing
3 = clearly Shakespearean style throughout
4 = strong Elizabethan English with authentic idioms and syntax
5 = authentic Shakespearean English indistinguishable from original

Response:
{response}

Reply with only a single integer: 0, 1, 2, 3, 4, or 5.
\end{verbatim}
\end{small}

\paragraph{KL benchmark definition.}
For each prompt, the SFT model first generates a reference response
autoregressively. Each evaluated model is then teacher-forced on
$[\text{prompt};\,\text{reference response}]$, and we compute
token-level KL divergence over the full vocabulary (response tokens
only):
\[
\mathrm{KL}_t(P\|Q)=\sum_v P_t(v)\bigl(\log P_t(v)-\log Q_t(v)\bigr).
\]
Reported values are averages over response tokens and then over
prompts, for $\mathrm{KL}(\mathrm{SFT}\|\mathrm{LCDD})$,
$\mathrm{KL}(\mathrm{SFT}\|\mathrm{LCDD{+}Trig})$, and
$\mathrm{KL}(\mathrm{Base}\|\mathrm{LCDD{+}Trig})$. \rev{Every evaluation prompt that also appears in the SFT/LCDD training set or the trigger optimization set is excluded. For the safety task, KL is computed on 189 WildJailbreak harmful prompts; for Fixed Response, on 199 Alpaca prompts; Shakespeare KL prompts have no overlap with training or trigger data.}

\paragraph{Sampling and decode defaults.}
Main benchmarks use 200 evaluation prompts per condition, sampled
with held-out seeds. Generation uses deterministic greedy decoding;
maximum generation length is 128 tokens for Fixed Response and
Shakespeare evaluation and 512 tokens for safety evaluation.


\end{document}